\documentclass{article}

\usepackage{arxiv}

\usepackage[utf8]{inputenc} 
\usepackage[T1]{fontenc}    
\usepackage{hyperref}       
\usepackage{url}            
\usepackage{booktabs}       
\usepackage{amsfonts}       
\usepackage{nicefrac}       
\usepackage{microtype}      
\usepackage{lipsum}		
\usepackage{graphicx}
\usepackage[numbers]{natbib}
\usepackage{doi}

\usepackage{lmodern}
\usepackage{microtype}
\usepackage{amssymb}
\usepackage{mathtools}
\usepackage{graphicx}
\usepackage{tabularx,booktabs}
\usepackage{xcolor}
\usepackage{multirow}
\usepackage{lineno}
\usepackage{url}

\usepackage{caption}
\usepackage{subcaption}

\usepackage{algorithm,algpseudocode}
\algnewcommand{\LeftComment}[1]{\State{ \(\triangleright\) #1}}

\title{Embedded Planogram Compliance Control System}

\author{ \hspace{1mm}M. Erkin Y\"{u}cel\\
	R\&D\\
	Migros Ticaret A.\c{S}.\\
	Istanbul, Turkey \\
	\texttt{mehmety@migros.com.tr} \\
	\And
	\hspace{1mm}Serkan Topalo\u{g}lu \\
	Faculty of Engineering, Department of Electrical and Electronics Engineering\\
	Yeditepe University\\
	Istanbul, Turkey \\
	\texttt{serkan.topaloglu@yeditepe.edu.tr} \\	
	\And
	\hspace{1mm}Cem \"{U}nsalan \\
	Faculty of Engineering, Department of Electrical and Electronics Engineering\\
	Yeditepe University\\
	Istanbul, Turkey \\
	\texttt{unsalan@yeditepe.edu.tr} \\
}

\renewcommand{\shorttitle}{Embedded Planogram Compliance Control System}

\hypersetup{
pdftitle={Embedded Planogram Compliance Control System},
pdfsubject={cs.cv},
pdfauthor={M. Erkin Y\"{u}cel, Serkan Topalo\u{g}lu, Cem \"{U}nsalan},
pdfkeywords={embedded camera modules, computer vision at the edge, single board computers, object detection, planogram compliance control, solar energy harvesting, RF energy harvesting},
}

\begin{document}
\maketitle

\begin{abstract}
	Smart retail stores are becoming an integral part of our lives. Various computer vision and sensor-based systems collaborate to enable these complex and automated operations. Furthermore, the retail sector presents several open and challenging problems that could benefit from advanced pattern recognition and computer vision techniques. One such critical challenge is planogram compliance control. In this study, we propose a complete embedded system to tackle this issue. Our system consists of four key components as image acquisition and transfer via stand-alone embedded camera module, object detection via computer vision and deep learning methods working on single board computers, planogram compliance control method again working on single board computers, and energy harvesting and power management block to accompany the embedded camera modules. The image acquisition and transfer block is implemented on the ESP-EYE camera module. The object detection block is based on YOLOv5 as the deep learning method and local feature extraction. We implement these methods on Raspberry Pi 4, NVIDIA Jetson Orin Nano, and NVIDIA Jetson AGX Orin as single board computers. The planogram compliance control block utilizes sequence alignment through a modified Needleman-Wunsch algorithm. This block is also working along with the object detection block on the same single board computers. The energy harvesting and power management block consists of solar and RF energy harvesting modules with suitable battery pack for operation. We tested the proposed embedded planogram compliance control system on two different datasets to provide valuable insights on its strengths and weaknesses. The results show that our method achieves F1 scores of 0.997 and 1.0 in object detection and planogram compliance control blocks, respectively. Furthermore, we calculated that the complete embedded system can work in stand-alone form up to two years based on battery. This duration can be further extended with the integration of the proposed solar and RF energy harvesting options.
\end{abstract}


\section{Introduction}\label{section:Introduction}

Technological advancements have been transforming the retail sector. To be more specific, integration of sensor networks, internet of things, computer vision, machine learning, and data analysis is making smart retail stores increasingly common and efficient. These technologies are practically applied to automatize operations and enhance customer experience. A significant challenge in retail is planogram compliance control which can be addressed through these technologies. Planogram is the diagram showing placement of products in shelves of a retail store. A shelf is considered planogram compliant when all the products on it are in the right place and quantity. Planogram compliance, crucial for maintaining proper product placement and stock levels, is often undermined by human error or stock issues, leading to sales loss and penalties. Typically, around 70\% compliance is observed in stores and resetting planograms can boost sales by up to 7.8\% in two weeks~\citep{Shapiro}.

Traditionally, planogram compliance is manually checked by employees, a process prone to labor intensity and errors. Alternatives include using mobile devices for photo-based checks~\citep{planorama}, installing fixed IP cameras~\citep{shelfie}, or employing robots~\citep{simbe, fellowrobots, badger}. However, these methods can be costly or laborious. Recently, embedded systems have been used to capture shelf images for efficient planogram control~\citep{focal, pricer, sesimotag}. In literature, various computer vision and pattern recognition methods address planogram compliance control and object detection in shelf images, as summarized in our previous study~\citep{Yucel}. There, we introduced a novel approach for planogram compliance control in smart retail, utilizing object detection using local feature extraction and implicit shape model (ISM) and iterative search processes, with sequence alignment via a modified Needleman-Wunsch algorithm. Melek~\emph{et al.}~\citep{Melek} developed a multi-stage system for grocery shelf product recognition, combining detection, classification, and refinement. 

\begin{figure*}[htbp]
	\centering
	\includegraphics[width=1\textwidth]{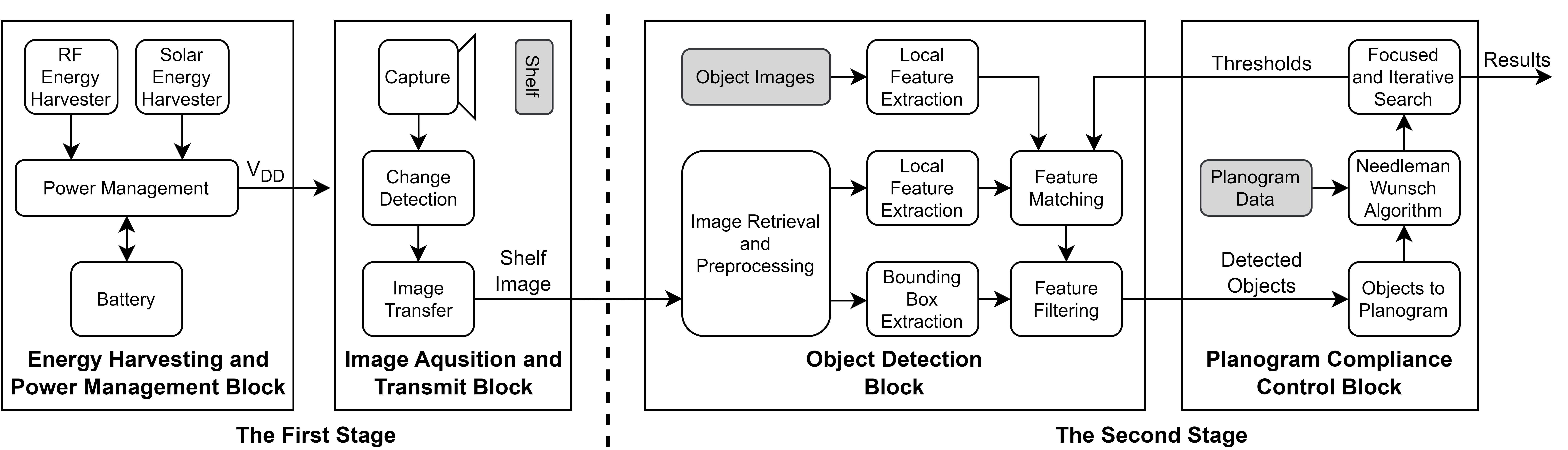}	
	\caption{Functional block diagram of the proposed system.}\label{fig:system}
\end{figure*}

In this study, we extend our previous work on planogram compliance control in two different ways. First, we propose using YOLO for bounding box detection and feature filtering, replacing ISM for object detection. Second, we form a complete embedded planogram compliance control system with all its components. While forming the system, we followed design constraints as follows. The system should operate stand-alone in a retail environment over extended periods, without needing for shelf cabling due to safety considerations. Given multiple shelves within a retail store, the system should also be cost-effective. High-resolution cameras, and high-end computer systems or cloud computing resources are impractical for this application due to their high power consumption and high cost, thus necessitating the use of embedded systems. The planogram compliance control system can work twice a day. For example, the first control can be made in the morning before the store opens. Hence, the employee can reorganize the shelves if compliance is below the desired level. To handle all these constraints, we propose an embedded system summarized as in Fig.~\ref{fig:system}.   

As can be seen in Fig.~\ref{fig:system}, the first stage of the system is the \emph{image acquisition and transfer} block formed by an embedded camera module with Wi-Fi connectivity. The camera to be used for this purpose should be low cost, low-power, and easy to acquire. The best fit for these constraints is the ESP-EYE module from Espressif Systems~\citep{espeye}. In the image acquisition and transfer block, the system captures shelf images at predetermined time intervals and transmits them to the second stage if it detects a significant change compared to the previous captured image. We will explain the details of this block in Section~\ref{sec:imageaqusition}. The image acquisition and transfer block is positioned in front of the shelf to work in stand alone form. Therefore, there should be an accompanying \emph{energy harvesting and power management} block to supply power to it. Moreover, the energy harvesting and power management block should have RF and solar energy harvesting modules to enhance the battery life. We will explain the working principles of this block in detail in Section~\ref{section:energyharvesting}. 

The second stage of the embedded system is formed by a single-board computer (SBC) to implement the \emph{object detection} and \emph{planogram compliance control} blocks. The SBC offers a cost-effective alternative to conventional PCs and can be distributed throughout the store for increased efficiency. In this study, we test Raspberry Pi 4, NVIDIA Jetson Orin Nano, and NVIDIA Jetson AGX Orin SBCs. The object detection block works such that it preprocess the shelf image and employs YOLO based bounding box detection and local feature extraction frameworks upon receipt an image from the first stage. In the planogram compliance control block, detected objects are converted into a planogram-compatible format. Then, the modified Needleman-Wunsch algorithm is applied for planogram compliance control of the shelf image. Furthermore, the focused and iterative search technique is used for improved results~\citep{Yucel}. We will explain the usage scenarios for object detection and planogram compliance control blocks in Sections~\ref{sec:objectdetection} and~\ref{sec:planogramcheck}, respectively. 

We will explain working principles of the proposed embedded planogram compliance control system in the following sections in detail. We test the proposed methods on two different datasets for the object detection and planogram compliance control steps. We will also analyze the overall system power consumption via tests. Furthermore, we test our methods on PC for benchmark.

\section{The Image Acquisition and Transfer Block}\label{sec:imageaqusition}

The image acquisition and transfer block is in the first stage of the proposed embedded planogram compliance control system. We will start with explaining its hardware properties in this section. Then, we will provide details on change detection and controlled image transfer operations performed on this hardware.

\subsection{Hardware used in Operation}

The camera to be used for the image acquisition and transfer block should be low cost and power. The best fit for these constraints is the ESP-EYE module from Espressif Systems~\citep{espeye}. This module incorporates an embedded camera and Wi-Fi chip. The camera can capture up to 2-Megapixel resolution images and store them in the external memory of the module. Therefore, the ESP-EYE module serves as a good candidate to implement and realize the image acquisition and transfer block in our embedded system. We use the C language to program the ESP-EYE module, benefiting from the Arduino IDE and ESP32 library.

A convenient retail store with 1000~$m^2$ sales area can have shelf length up to 1200~m. An ESP-EYE can cover an average of 4~m wide shelf blocks with three racks. Hence, it can cover a total of 12~m shelf area. As a result, we will need approximately 100 of ESP-EYE modules to cover all the shelves in a typical retail store.

\subsection{Change Detection and Controlled Image Transfer}

Planogram compliance control is necessary only when there is a major change in the shelf. Therefore, we pick the image rationing with preprocessing method from our previous work~\citep{Yucel2}. In this method, we first acquire the shelf image, $I_l$, in grayscale format with QVGA resolution and then apply filtering to obtain $\tilde{I_l} = G\ast I_l$, where $G$ is the simple Gaussian blurring filter kernel. The previously captured and filtered image, $\tilde{I_r}$, is stored in flash memory of the ESP-EYE module for reference during low power hibernation mode. Subsequently, we calculate ratio of the images, where each pixel ratio approaches zero with no change and nears $\pi/4$ with maximum change. After thresholding each pixel ratio with a predetermined value $\tau_p$ and counting the changed pixels, we determine if the shelf image has changed based on a predefined threshold, $\tau_c$, of all pixels. If any change is detected in a shelf image, a new JPEG image with UXGA resolution $I$ is acquired and transmitted to a remote location via Wi-Fi. Finally, the ESP-EYE module enters low power hibernation mode, awaiting the next timer trigger to wake up. We provide the pseudocode for the overall operation in Algorithm~\ref{alg:imageaqusition}.

\begin{algorithm}[htbp]
	\caption{Pseudocode of image capture and rationing for shelf control.}\label{alg:imageaqusition}
	\begin{algorithmic}
		\State{$I_l \gets \Call{CaptureImage}$}
		\State{$\tilde{I_l} \gets \Call{FilterImage}{I_l}$}
		\State{$c \gets \Call{DetectChange}{\tilde{I_r}, \tilde{I_l}, \tau_p}$}
		\If{$c \geq \tau_c$}
			\State{$I \gets \Call{CaptureImage}$}
			\State{$\Call{EnableWiFi}$}
			\State{$\Call{TransferImage}{I}$}
			\State{$\Call{DisableWiFi}$}
		\EndIf{}
		\State{$\Call{EnterHibernationMode}$}
	\end{algorithmic}
\end{algorithm}

\section{The Object Detection Block}\label{sec:objectdetection}

When any change in the shelf image is detected, we should find objects in the shelf image to compare them with the actual planogram. We will explain the operations needed for the object detection step in this section. We will start with the hardware used in operation next.

\subsection{Hardware used in Operation}

The object detection block in our embedded planogram compliance system requires high processing power and memory. Therefore, we deploy an embedded SBC to implement it. To be more specific, we use SBCs with high processing power as Raspberry Pi 4, NVIDIA Jetson Orin Nano, and NVIDIA Jetson AGX Orin to find objects in the image. We utilize advanced software and libraries on these embedded devices, leveraging their capability to run the Linux OS\@. We use OpenCV library for local feature extraction and object detection operations. We picked NodeJS as the development framework at SBCs. Furthermore, we compile the OpenCV from its source to deploy the DNN and image processing functions in the NodeJS environment. We enabled the CUDA support for OpenCV while compiling it in NVIDIA Jetson devices which gives us to use their GPU parallel processing capability for DNN operations.

\subsection{Image Retrieval and Preprocessing}\label{sec:imageretrieval}

As the image acquisition and transfer block sends the shelf image to the object detection block, an end-point should receive it. We implement this operation using Express which is NodeJS web application framework~\citep{Express}. We create a web server application which listens for the incoming image data and upon receive, it saves the image data to a file. 

The received image contains full shelf view containing several shelf racks. However, the planogram compliance control block works with a single shelf rack. Thus, we split the image horizontally into the number of shelf racks using OpenCV library functions. Depending on the distance of the camera to the shelf, there are three to four shelf racks in the shelf view. Hence, a shelf rack image approximately has dimensions as $1600 \times 400$ pixels after splitting. We also add padding to the top and bottom of the image. Hence, it becomes suitable to be processed in the bounding box step via YOLO\@. Finally, we employ a blurring filter to the image using OpenCV library functions to remove noise.

\subsection{Local Feature Extraction and Matching}\label{sec:localfeatureextraction}

Objects on shelves often display complex and closely resembling appearances in the retail environment. This study employs local feature extraction, specifically the SIFT method for object detection, a technique chosen based on its superior performance in our previous study~\citep{Yucel}. Let $I(x, y)$ represent a shelf image with $J$ different objects in it. Assume that the $j^{th}$ object on the shelf has a model (or planogram) image $I_j(x, y)$ for $j = 1, \ldots, J$. Let $w_j$ and $h_j$ be the width and height of $I_j(x, y)$, respectively. We extract local feature vectors, $\overrightarrow{f_{jl}}$ for $l = 1, \ldots, L_j$ from $I_j(x, y)$ to represent the $j^{th}$ object with $L_j$ keypoints, using SIFT functions in OpenCV library. Similarly, we extract the local feature vectors, $\overrightarrow{f_m}$  for $m = 1, \ldots, M$ from the shelf image, $I(x, y)$, to detect objects in it. Here, $M$ is the total number of extracted keypoints from the shelf image. We provide the pseudocode for the feature extraction operation in Algorithm~\ref{alg:localfeatextract}. 

\begin{algorithm}[htbp]
	\caption{Pseudocode of local feature extraction function.}\label{alg:localfeatextract}
	\begin{algorithmic}
		\Function{ExtractFeat}{$image$}
			\State{$kp \gets \Call{SiftKeypointDetect}{image}$}
			\State{$f \gets \Call{SiftFeatureDescriptor}{image, kp}$}
			\State{\textbf{return} $f$}
		\EndFunction{}
	\end{algorithmic}
\end{algorithm}

One way of finding the object of interest in the shelf image is matching extracted local feature vectors from the $j^{th}$ object, $\overrightarrow{f_{jl}}$, and shelf image, $\overrightarrow{f_m}$. Here, we use the brute force matching functions in OpenCV library to find the matching features. We provide the pseudocode for this operation in Algorithm~\ref{alg:bfmatch}. 

\begin{algorithm}[htbp]
	\caption{Pseudocode of feature matching function.}\label{alg:bfmatch}
	\begin{algorithmic}
		\Function{FeatMatch}{$f_{jl}, f_m, \tau_\alpha$}
			\State{$matches \gets \Call{BFMatcher}{f_{jl}, f_m}$}
			\State{$f_{jn} \gets \{\}$}
			\ForAll{$m \in matches$}
				\If{$m[0].d < \tau_\alpha \times m[1].d$}
					\State{$f_{jn} \gets f_{jn} \bigcup m$}
				\EndIf{}
			\EndFor{}
			\State{\textbf{return} $f_{jn}$}
		\EndFunction{}
	\end{algorithmic}
\end{algorithm}

Furthermore, we also use the ratio test proposed by Lowe~\citep{Lowe} to keep only strong matches. Lowe suggests setting $\tau=0.8$ gives the best result for SIFT\@. Instead of such a constant value, we set the matching threshold as $\tau_\alpha = 0.95 - 0.2\alpha$ where $\alpha$ is the iteration parameter (initially set to one) to be introduced in~\ref{sec:focusedanditerativesearch}. Assume that we have $N_j$ local feature vectors from $I(x,y)$ satisfying the threshold constraint for the $j^{th}$ object. Hence, we will have $\left\{\overrightarrow{f_{jn}}\right\} \subset \left\{\overrightarrow{f_m}\right\} $ for $n = 1, \ldots, N_j$ and $N_j \leq  M$.

\subsection{Bounding Box Extraction and Filtering}\label{sec:boundingboxextraction}

YOLO (You Only Look Once) introduced by Redmon~\emph{et al.}~\citep{Redmon} is an advanced, real-time object detection system, known for its efficiency and speed. It integrates bounding box prediction and object classification into a single process. Unlike traditional methods, YOLO divides the input image into a grid, with each cell predicting bounding boxes and class probabilities, enabling much faster processing speeds. YOLO has different versions where each version builds upon the foundational principles of its predecessors, offering enhanced performance in terms of training speed and prediction accuracy. 

\begin{figure*}[htbp]
	\centering
	\begin{subfigure}{0.9\textwidth}
		\includegraphics[width=1\textwidth]{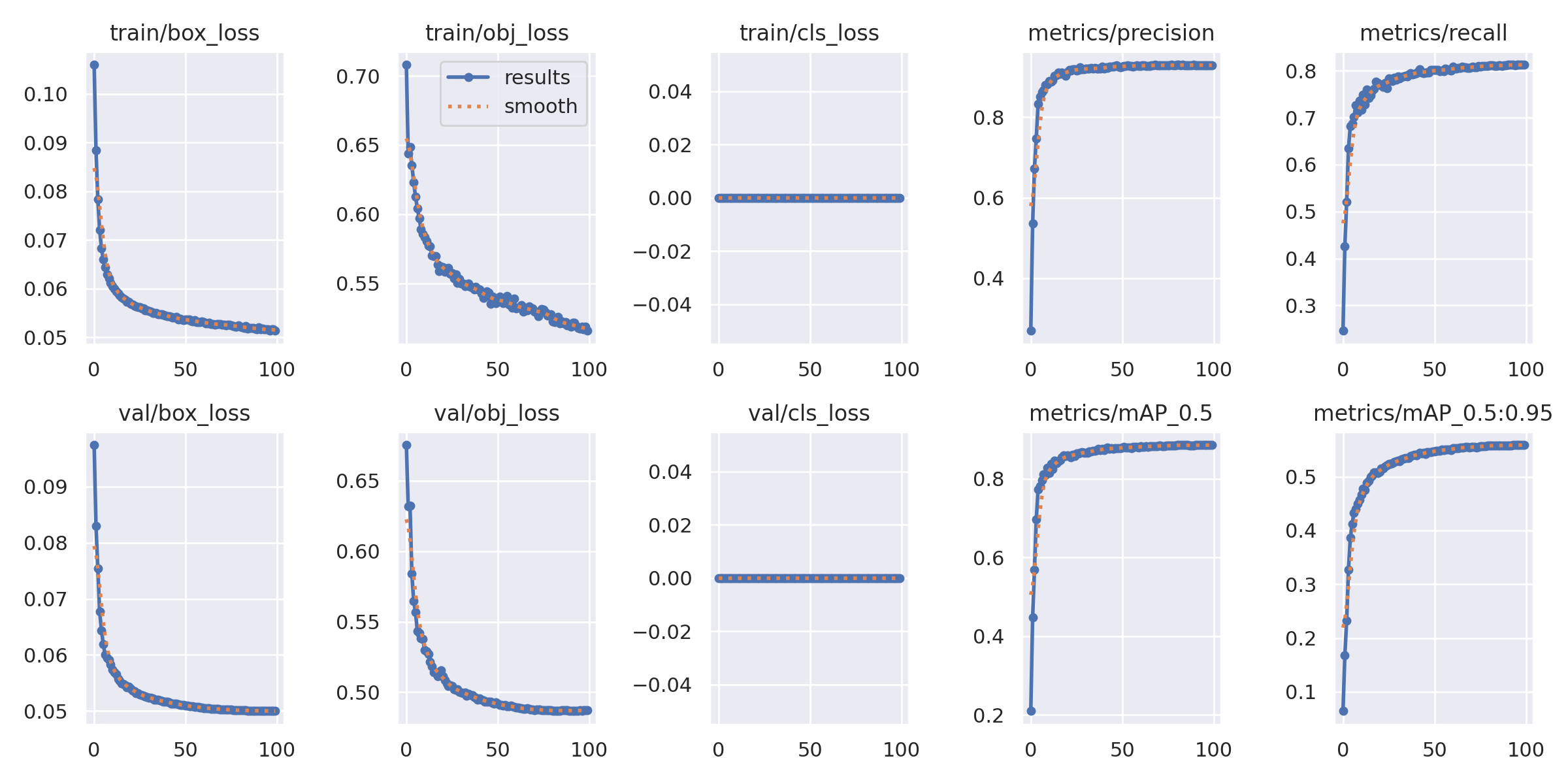}\caption{YOLOv5s}\label{fig:yolov5sres}
	\end{subfigure}\\
	\begin{subfigure}{0.9\textwidth}
		\includegraphics[width=1\textwidth]{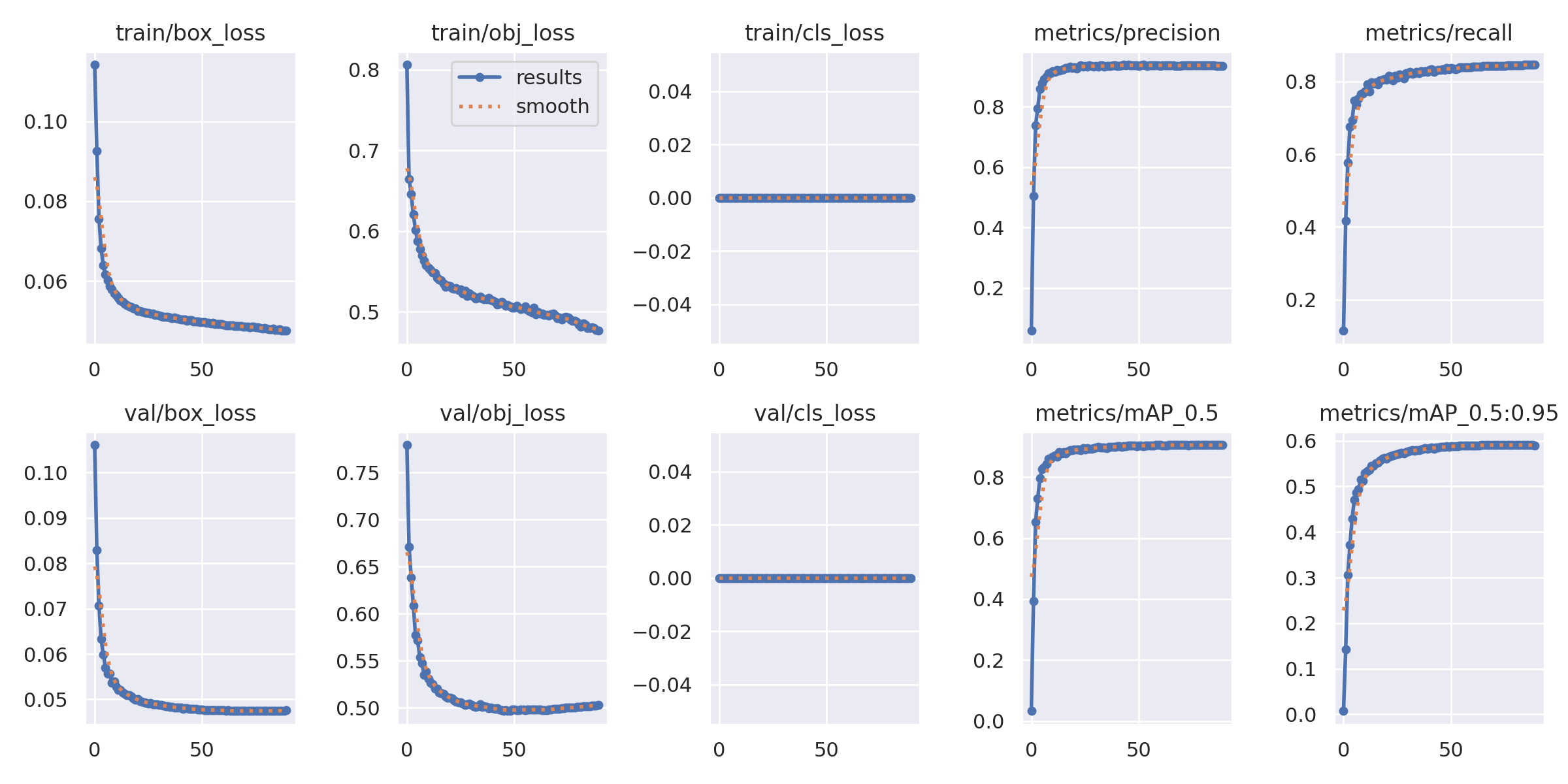}\caption{YOLOv5s}\label{fig:yolov5xres}
	\end{subfigure}
	\caption{YOLOv5 training results.}\label{fig:YOLOtraining}
\end{figure*}

One of the most popular YOLO version is YOLOv5~\citep{yolov5} which represents a significant leap in terms of performance and usability. This version incorporates several state-of-the-art techniques, such as the use of Cross Stage Partial networks (CSPNet) for enhancing the learning capability of the model, and the application of Spatial Pyramid Pooling (SPP) layers to increase the receptive field. Additionally, YOLOv5 is designed to be user-friendly, with simpler model training, easier customization, and better support for deployment across various platforms, including embedded devices. YOLOv5 has ten available multiscale models having speed/accuracy trade-offs. It supports 11 different formats (both export and run time) and can be easily implemented in embedded devices. Hence, in this study we choose YOLOv5 as our object detection model, specifically YOLOv5s and YOLOv5x models. We did not use pre-trained models. We trained them on the SKU110K dataset~\citep{Goldman}, which has single class, using Nvidia Jetson AGX\@. We set the image size to $640\times640$ and epoch length to 100 in the training. We also set the batch size to 48 and 144 for YOLOv5x and YOLOv5s, respectively. The training lasts 16.6 hours and the mAP50 reaches 0.59 for YOLOv5x, and it lasts 11.4 hours and the mAP50 reaches to 0.56 for YOLOv5s, respectively. We give the graphs for training and validation in Fig.~\ref{fig:YOLOtraining}.


We employ the OpenCV DNN library to perform inference on shelf images using the YOLOv5 model. OpenCV DNN library cannot use the YOLOv5 model file directly. However, it can use ONNX format for model file. Hence, we first export our custom trained YOLOv5 model to the ONNX format. The model produces all the candidate bounding boxes as a result of the inference operation. Each bounding box, $B_b$, is characterized by a tuple containing the confidence score and spatial dimensions as $(cs_b, x_b, y_b, w_b, h_b)$ for $b = 1, \ldots, B$. To identify the best candidates for the $j^{th}$ object, we apply filtering and threshold techniques. Specifically, we use the width $w_j$, height $h_j$, and aspect ratio $\frac{w_j}{h_j}$ of the $j^{th}$ object as reference. We discard any bounding box with dimensions and aspect ratio falling below half or exceeding twice these reference values. Subsequently, we apply a confidence score threshold, which we set it at 5\% of the maximum confidence score, to further refine our candidates. Hence, we discard any bounding box with very low confidence score. Following these steps, we obtain $C_j$ candidate bounding boxes for the $j^{th}$ object. We provide the pseudocode explaining these steps in Algorithm~\ref{alg:boundingboxextract}.

\begin{algorithm}[htbp]
	\caption{Pseudocode of bounding box extraction and filtering.}\label{alg:boundingboxextract}
	\begin{algorithmic}
		\Function{ExtractBB}{$I$}
			\State{$B_b \gets \Call{Inference}{I}$}
			\ForAll{$Object_j \in planogram$}

				\State{$B_{tmp} \gets \{\}$}
				\ForAll{$box_b \in B_b$}
					\If{$(w_j \times 0.5 \leq w_b \leq w_j \times 2)$ and 
					\State{$(h_j \times 0.5 \leq h_b \leq h_j \times 2)$ and }
					\State{$(\frac{w_j}{h_j} \times 0.5 \leq \frac{w_b}{h_b} \leq \frac{w_j}{h_j} \times 2)$}}
							\State{$B_{tmp} \gets B_{tmp} \cup \{box_b\}$}
					\EndIf{}
				\EndFor{}	

				\State{$\tau_b \gets 0.05 \times \max(\forall cs_{tmp})$}
				\State{$B_{jc} \gets \{\}$}
				\ForAll{$box_{tmp} \in B_{tmp}$}
					\If{$(cs_{tmp} \geq \tau_b)$}
						\State{$tl \gets (x_{tmp} - \frac{w_{tmp}}{2}, y_{tmp} - \frac{h_{tmp}}{2})$}
						\State{$br \gets (x_{tmp} + \frac{w_{tmp}}{2}, y_{tmp} + \frac{h_{tmp}}{2})$}
						\State{$cnt \gets (x_{tmp}, y_{tmp} )$}
						\State{$B_{jc} \gets B_{jc} \cup  \{ tl, br, cnt, cs_{tmp} \} $}
					\EndIf{}
				\EndFor{}
			\EndFor{}
			\State{\textbf{return} $B_{jc}$}
		\EndFunction{}	
	\end{algorithmic}
\end{algorithm}

As we detect possible object bounding boxes for the $j^{th}$ object in the shelf image, the next step is finding matching local features inside bounding boxes. Therefore, we find the number of features inside each bounding box, $s_c$. We further eliminate weak objects which have lower feature number than 5\% of the number of features of the $j^{th}$ object, $L_j$. Then, we normalize it with the total number of feature of the $j^{th}$ object, $L_j$. We multiply it with the confidence score of the bounding box to use it as the detection score for the detected object. Some bounding boxes may overlap, meaning that there may be more than one candidate object in the given location. Therefore, we eliminate weak candidates using the greedy algorithm for non-maxima suppression (NMS) of bounding boxes. At the end of this operation, we detect a total of $D$ objects, $D \leq \sum_{j = 1}^{J} C_{j}$, from the shelf image $I(x, y)$. We provide the pseudocode to filter out matching features not lying in a bounding box in Algorithm~\ref{alg:featurefilter}. 

\begin{algorithm}[htbp]
	\caption{Pseudocode of matching feature filtering.}\label{alg:featurefilter}
	\begin{algorithmic}		
		\Function{FeatFilter}{$B_{jd}, f_{jn}$}
			\ForAll{$Object_j \in planogram$}
				\State{$B_{tmp} \gets \{\}$}
				\ForAll{$box_c \in B_{jc}$}				

					\State{$s_c \gets 0$}
					\ForAll{$feature_n \in f_{jn}$}					
						\If{$(tlx_{c}  \leq x_n \leq brx_{c} )$ and 
						\State{$(tly_{c}  \leq y_n \leq bry_{c} )$ }}
								\State{$s_c \gets s_c + 1$}
						\EndIf{}
					\EndFor{}

					\If{$s_c > 0.05 \times L_j $}
						\State{$weight_c \gets \frac{s_c}{L_j} \times cs_c$}
						\State{$B_{tmp} \gets B_{tmp} \cup \{box_c, weight_c\}$}
					\EndIf{}
				\EndFor{}
				\State{$D_{jd} \gets \Call{NMS}{B_{tmp}}$}
			\EndFor{}
			\State{\textbf{return} $D_{jd}$}
		\EndFunction{}
	\end{algorithmic}
\end{algorithm}

\section{The Planogram Compliance Control Block}\label{sec:planogramcheck}

Object detection results in Section~\ref{sec:objectdetection} lead to planogram compliance control. After representing detection results in an abstract planogram format, we use the modified Needleman-Wunsch algorithm to align and compare the detected and reference planograms as suggested in our previous study~\citep{Yucel}. Then, we use the iterative search method to focus on undetected or empty object locations in the shelf image. We explain these steps in detail next. Here, we use the same hardware used in Section~\ref{sec:objectdetection} as well as the NodeJS framework. 

\subsection{Sequence Alignment}

To compare the detected objects in previous section with the given planogram data, we first need to convert detected object list into planogram format. Here, we sort the detected objects from left to right according to their center coordinates. Then, we group the items of same object hence we can compare their location and quantity, later. We also keep the start and end coordinates hence we can iteratively search for undetected objects only as explained in our previous study~\citep{Yucel}. Based on this, we represent the detected objects in the shelf image (in sorted and grouped form) as $L_s = \left[ o_d, q_d, B_d\right]$ and reference planogram as $L_r = \left[ o_t, q_t, B_t\right]$. To do so, we use the algorithm given in Algorithm~\ref{alg:obj2pla}. 

\begin{algorithm}[htbp]
	\caption{Pseudocode of object to planogram method.}\label{alg:obj2pla}
	\begin{algorithmic}						
		\Function{Obj2Pla}{$D_{m}$}			
			\State{$L \gets \{\}$}

			\State{$D_{m} \gets \Call{Sort}{D_{m}}$}
			\ForAll{$o_m\in D_{m}$}
				\If{$L \owns o_m$}
					\State{$q_m \gets q_m + 1$}
					\State{$br \gets br_m$}
				\Else{}				
					\State{$q_m  \gets 1$}
					\State{$tl \gets tl_m$, $br \gets br_m$}
				\EndIf{}

				\State{$B_m  \gets \{ tl, br\} $}
				\State{$L  \gets L \cup \{ o_d, q_d, B_d \}$}
			\EndFor{}		

			\State{\textbf{return} $L$}
		\EndFunction{}
	\end{algorithmic}
\end{algorithm}


We can control planogram compliance by comparing the reference, $L_r$, and detected, $L_s$, planograms for a given shelf. To do so, we need to align them using the modified Needleman-Wunsch (NW) algorithm~\citep{Needleman}. The NW algorithm has three steps: initialization, matrix fill and trace back. In the initialization step we set the score matrix $F$ with size $(E+1)\times(T+1)$ and set the first row and column of $F$. In the matrix filling step we fill the $F(d,t)$ using the gap penalties, $g_{ins}$ and $g_{del}$, for insert and delete, and the substitution score, $s(o_d, o_t)$, for match and mismatch. In this study, we set the gap penalties and substitution score dynamically based on their importance. We set the gap penalty for the delete operation as the required number of objects in the reference planogram since we could not detect them. We set the gap penalty for the insert operation as the number of extra items in the detected planogram. The aim here is that the more inserted such false objects are, the more penalty there should be. The third step in the modified NW algorithm is tracing back the filled matrix entries. Here, we trace back the decisions we have made while calculating each matrix entry. We use the pseudocode in Algorithm~\ref{alg:nw} for planogram alignment using the modified NW algorithm.

\begin{algorithm}[htbp]
	\caption{Pseudocode of the Needleman-Wunsch algorithm.}\label{alg:nw}
	\begin{algorithmic}		
		\Function{NW}{$L_r, L_s$}

		\State{\LeftComment{Initialize}}
		\State{$F \gets [E+1][T+1]$}
		\State{$F_{trace} \gets [E+1][T+1]$}
		\State{$t \gets 1, d \gets 1$}
		\State{$F(0,0) \gets 0$}
		\While{$t \leq T $}
			\State{$F(0,t) \gets F(0,t-1) - 1$}
			\State{$t \gets t + 1$}
		\EndWhile{}
		\While{$d \leq E $}
			\State{$F(d, 0) \gets F(d-1,0) - 1$}			
			\State{$d \gets d + 1$}
		\EndWhile{}

		\State{\LeftComment{Matrix Fill}}
		\ForAll{$cell(d,t) \in F$}
			\State{$up \gets F(d, t - 1) - g_{del}$}
			\State{$left \gets F(d - 1, t) - g_{ins}$}
			\State{$dia \gets F(d - 1, t - 1) + s(o_d, o_t)$}

			\State{$F(d,t) \gets \max(up, left, dia)$}
			\State{$F_{trace}(d,t)  \gets maxIndex(up, left, dia)$}
		\EndFor{}
					
		\State{\LeftComment{Trace Back}}
		\State{$d \gets E+1, t \gets T+1$}
		\While{$(d,t) \neq (0,0)$}
			\If{$F_{trace}(d,t) = dia$}
				\State{$\hat{L_r} \gets \hat{L_r} \bigcup \left[ o_t, q_t, B_t\right]$}
				\State{$\hat{L_s} \gets \hat{L_s} \bigcup \left[ o_d, q_d, B_d\right]$}
				\State{$(d,t) \gets (d-1, d-1)$}
			\ElsIf{$F_{trace}(d,t) = up$}
				\State{$\hat{L_r} \gets \hat{L_r} \bigcup \left[ A, 0, -\right]$}
				\State{$\hat{L_s} \gets \hat{L_s} \bigcup \left[ o_d, q_d, B_d\right]$}
				\State{$(d,t) \gets (d, t-1)$}
			\ElsIf{$F_{trace}(d,t) = left$}
				\State{$\hat{L_r} \gets \hat{L_r} \bigcup \left[ o_t, q_t, B_t\right]$}
				\State{$\hat{L_s} \gets \hat{L_s} \bigcup \left[ D, 0, -\right]$}
				\State{$(d,t) \gets (d-1, t)$}
			\EndIf{}
		\EndWhile{}
			\State{\textbf{return} $\hat{L_r}, \hat{L_s}$}
		\EndFunction{}
	\end{algorithmic}
\end{algorithm}

Following the steps in Algorithm~\ref{alg:nw}, we can obtain aligned object groups $\hat{o}_d$ and $\hat{o}_t$ and their quantities $\hat{q}_d$ and $\hat{q}_t$. We then calculate the planogram match ratio, $\mu_s$, ranging from 0 (no compliance) to 1 (full compliance), using the Algorithm in Algorithm~\ref{alg:placontrol}. A $\mu_s$ of 0 means all items are misplaced, while 1 signifies complete planogram compliance. A match ratio between 0 and 1 indicates partial compliance, with some items missing, extra, or incorrectly placed. 

\begin{algorithm}[htbp]
	\caption{Pseudocode of planogram control operations.}\label{alg:placontrol}
	\begin{algorithmic}	
		\Function{PLAControl}{$\hat{L_r}, \hat{L_s}$}
			\State{$result \gets \{\}$}
			\State{$sum \gets 0, csum \gets 0$}
			\ForAll{$\hat{d} \in \hat{L_s}$ and $\hat{t} \in \hat{L_r} $}
				\If{$o_d == o_t$}
					\If{$q_d == q_t$}
						\State{$result \gets result \bigcup MT$}
						\State{$sum \gets sum + q_d$}
					\ElsIf{$q_d > q_t$}
						\State{$result \gets result \bigcup ME$}
						\State{$sum \gets sum + \min(q_d, q_t)$}
					\ElsIf{$q_d < q_t$}
						\State{$result \gets result \bigcup MI$}
						\State{$sum \gets sum + \min(q_d, q_t)$}
					\EndIf{}
				\ElsIf{$o_d \neq o_t$}
					\State{$result \gets result \bigcup NM$}
				\EndIf{}
				\State{$csum \gets csum + q_t$}
			\EndFor{}
			\State{$\mu_s \gets \frac{sum}{csum}$}
			\State{\textbf{return} $result, \mu_s$}
		\EndFunction{}
	\end{algorithmic}
\end{algorithm}

\subsection{Focused and Iterative Search}\label{sec:focusedanditerativesearch}

The object detection method and modified NW algorithm introduced in the previous sections may not produce exact planogram alignment result in one iteration. Therefore, we use the focused and iterative search method to improve the results~\citep{Yucel}. Here, we repeat the object detection operation by relaxing the detection constraints and focusing on the region of interest (ROI) in the shelf image $I(x,y)$. As we form the ROI, we iteratively decrease the value of $\alpha$ as $\alpha_{new} = 0.75 \alpha_{old}$. We detect new objects based on the new $\alpha$ value at each iteration and apply the NW algorithm to the updated $L_s$ and $L_r$ until $\mu = 1$ or $\mu$ does not change for six successive iterations. We provide the pseudocode for the focused and iterative search operation in Algorithm~\ref{alg:fis}. As the iterations end, we have the planogram compliance control result at hand.

\begin{algorithm}[htbp]
	\caption{Pseudocode of focused and iterative search operations.}\label{alg:fis}
	\begin{algorithmic}	
		\Function{FocusedItrSearch}{$I,  L_r$}
			\State{$\alpha \gets 1, retryNum \gets 0$}
			\State{$\mu_s \gets 0, \mu_{sp} \gets 0$}
			
			\State{$ \overrightarrow{f_m} \gets \Call{ExtractFeat}{I}$}
			\State{$ B_{jd} \gets \Call{ExtractBB}{I}$}

			\While{$IterNum < 6$ and $\mu_s \neq 1$}
				\State{$\mu_{sp} \gets \mu_s$}
				\State{$\tau_\alpha \gets 1 - 0.15\alpha $}
				\ForAll{$object_j \neq MT \in result$}
					\State{$f_{jn} \gets \Call{FeatMatch}{f_{jl}, f_m, \tau_\alpha}$}
					\State{$D_{jd} \gets \Call{FeatFilter}{f_{jl}, f_{jn}}$}
				\EndFor{}
				
				\State{$L_s \gets \Call{Obj2Pla}{D_{jd}}$}
				\State{$\hat{L_r}, \hat{L_s} \gets \Call{NW}{L_r, L_s}$}
				\State{$result, \mu_s \gets \Call{PLAControl}{\hat{L_r}, \hat{L_s}}$}

				\If{$\mu_{sp} == \mu_s$}
					\State{$retryNum \gets retryNum + 1$}
				\EndIf{}
			\EndWhile{}
		
			\State{\textbf{return} $result, \mu_s$}
		\EndFunction{}
	\end{algorithmic}
\end{algorithm}

\section{The Energy Harvesting and Power Management Block}\label{section:energyharvesting}

We need a battery-based, stand-alone system for the image acquisition and transfer block. Therefore, one important part of the proposed embedded planogram compliance control system is the energy harvesting and power management block. Various energy harvesting methods, including RF and WPT, are increasingly utilized to extend the battery life of embedded systems~\citep{Riva,Choi,Tripathi,Warnakulasuriya}. Kadir~\emph{et al.}~\citep{kadir} introduced a Wi-Fi energy harvester suitable for indoor, low-power applications, featuring multiple antennas and providing up to 2~V output. Tepper~\emph{et al.}~\citep{Tepper} explored the use of WPT systems in aircraft for powering sensors, employing commercial Powercast 2110 evalkit with a transmitter operating at 915~MHz. 

In this section, we explain suitable modules for solar and RF energy harvesting. We use them in our embedded planogram compliance control system to extend battery life, either separately or combined using complementary methods or power ORing~\citep{Estrada}. Therefore, we start with explaining the battery pack used in operation.

\subsection{Battery Pack}

The battery generally has the largest size in an embedded system with its size directly proportional to its capacity and price. Hence, determining the battery life and capacity plays a vital role while determining the system physical dimensions and overall cost. We maximize the battery life with the help of ambient energy harvesting options in the proposed system. Therefore, we should use a rechargeable battery to store harvested energy. To be more specific, we picked the LiFePO4 battery, which is a type of Lithium-Ion (Li-Ion) battery, to supply the system since its voltage range is suitable directly use it with ESP-EYE module. Moreover, we minimize the current drawn via utilizing hibernation mode for the ESP-EYE and periodic wake-ups via timer interrupts. This operation significantly saves battery power and maximizes the battery life.

\subsection{Solar Energy Harvesting}

The simplest energy harvesting method is to use ambient light with the help of photovoltaic (PV) cells, as there is usually constant artificial lighting in a retail store. We use a power management integrated circuit (PMIC) to charge the battery from the harvested energy, track the maximum power point of PV cells, and supply the regulated voltage to the system. We use the SPV1050 from STMicroelectronics as the PMIC that can be used for indoor light energy harvesting~\citep{spv1050}. This module harvests energy from solar cells and stores the harvested energy in a capacitor. Then, it supplies the regulated output voltage to system components. It can harvest energy from PV cells even if the output voltage of the PV cell is as low as 150~mV. It also supports secondary input where other energy harvesting methods can be connected in complementary to the PV cell. Hence, it can harvest energy from this input if there is not enough lighting on PV cell. De Rossi~\emph{et al.}~\citep{Derossi} show that an amorphous silicon (a-Si) PV cell can produce about 13~$\mu W/cm^2$ under 500~lux LED light illumination, which is the average illumination level in a retail store. Hence, we pick the a-Si PV cell in the proposed system. 

\subsection{RF Energy Harvesting}

We can also use the RF energy harvesting chip P2210B from Powercast to back up the battery power of ESP-EYE module. Basically, the system converts RF energy to DC voltage and stores it in a capacitor. Then, the chip regulates and supplies the voltage to output. The P2210B chip can harvest power level below to -12~dBm from 902~MHz to 928~MHz RF signals. It can provide an output current of 0.2~mA at an input power level of 0~dBm. The 0~dBm average RF power at indoor is quite acceptable when it is recognized that maximum power level is 36~dBm and reduced down to 4~dBm at 915~MHz for GSM communication~\citep{3GPPTS0505}. 

\section{Experiments}\label{section:Experiments}

We test the proposed system in this section. To do so, we benefit from two datasets. We first introduce them. Then, we provide the object detection performance of the proposed method on these datasets. To note here, we take the performance of the SIFT method in our previous work as the baseline method~\citep{Yucel}. Furthermore, we re-assess object detection performance the baseline method using intersection over union (IoU) value of 50\%. Afterward, we provide experimental results on the planogram compliance control step. We also provide timing and power consumption performance of the proposed embedded system. 

\subsection{Datasets used in Experiments}

In the absence of a dedicated dataset for evaluating embedded system object detection in retail, we utilized available datasets relevant to similar problems, such as object detection in retail stores and bounding box extraction. Specifically, we adapted a subset of the grocery product dataset annotated by Tonioni and Di Stefano~\citep{George, Tonioni}. We split the images for single shelf rack focus and resized them to 400 pixels in height, matching the dimensions used by the image retrieval block in our system. We also employed the Migros dataset, selecting its first subset, comprising 28 shelf images with 312 objects, already split for single rack views and resized to align with our system's requirements~\citep{Yucel}. This approach allows an effective assessment of our method and comparison with previous works.

\subsection{Object Detection Performance}

\begin{figure*}[htbp]
	\centering
	\begin{subfigure}{0.8\textwidth}
		\centering
		\includegraphics[width=\textwidth]{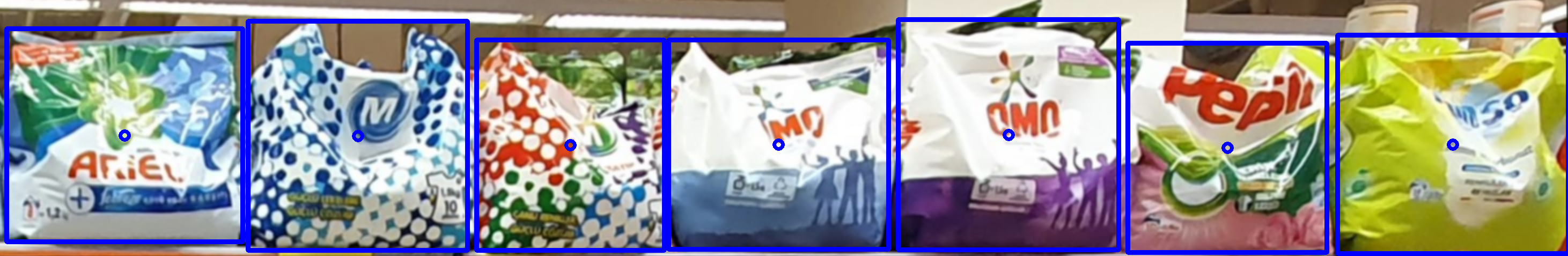}\caption{Migros dataset.}\label{fig:res_migros}
	\end{subfigure}\\
	\begin{subfigure}{0.8\textwidth}
		\centering
		\includegraphics[width=\textwidth]{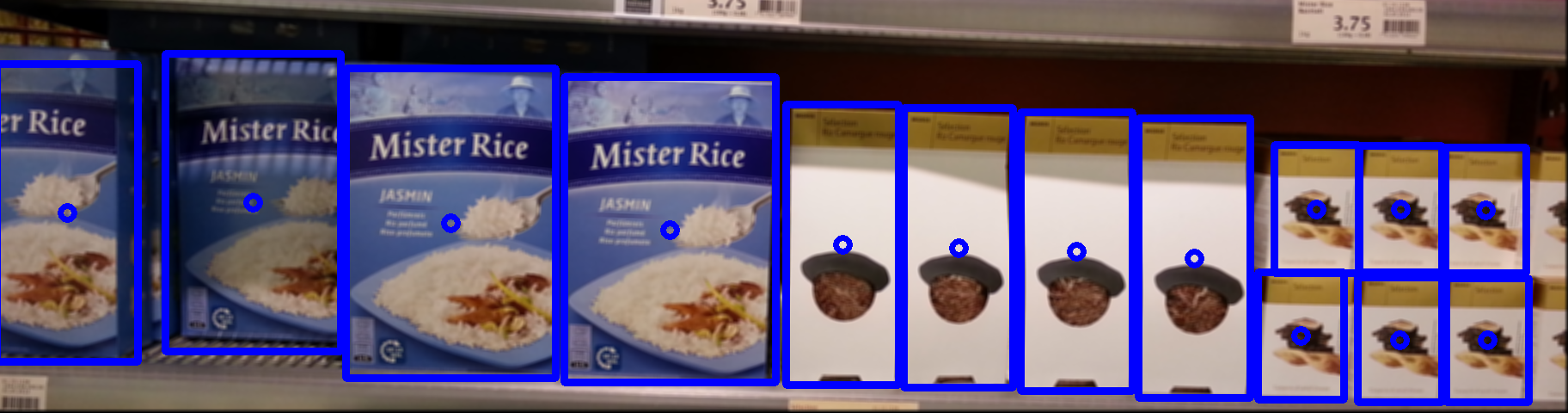}\caption{Grocery Product dataset.}\label{fig:res_grozi}
	\end{subfigure}
	\caption{Bounding box of the detected objects from the representative shelf images of test datasets.}\label{fig:res}
\end{figure*}

We focus on the object detection performance of the proposed method in this section. Here, we use IoU value of 50\% while computing the precision, recall, and F1 score as our performance metrics. We tested the proposed object detection method on the Migros and grocery product dataset. We provide a representative image from each dataset for the detected objects in Fig.~\ref{fig:res}. As can be seen in this figure, all objects are detected correctly with the proposed method. 

We provide object detection performance of the proposed method on the Migros dataset in Table~\ref{table:pla_obj_perf_test_migros}. As can be seen in this table, the proposed method in this study achieves higher object detection performance compared to the results of baseline object detection method. To be more specific, the overall performance increases using YOLO instead of ISM while detecting bounding boxes. To note here, bounding box detection and classification steps depend on the SIFT keypoint extraction and feature matching performance in the baseline method. We benefit from YOLO to detect bounding boxes in the proposed system. Therefore, we can reach F1 scores around 1 for both YOLOv5 models. This is a remarkable result for the given problem. 

\begin{table}[htbp]
	\centering
	\caption{Object detection performance of the proposed method on Migros dataset.}\label{table:pla_obj_perf_test_migros}
	\begin{tabular}{lccc}
		\toprule
        \textbf{Model}		& \textbf{Precision}	& \textbf{Recall}	& \textbf{F1 Score}		\\
		\midrule
		\textbf{Baseline}	& 0.974					& 0.965				& 0.969					\\
		\textbf{YOLOv5s}	& 0.997					& 0.997				& 0.997 				\\
		\textbf{YOLOv5x}	& 0.994					& 0.990				& 0.992 				\\
		\bottomrule
	\end{tabular}
\end{table}

We provide object detection results on the grocery product dataset in Table~\ref{table:pla_obj_perf_test_grozi}. As can be seen in this table, the proposed method with YOLOv5x model achieves higher object detection performance on this dataset compared to the baseline object detection method. However, overall performance of the YOLO models decrease compared to the Migros dataset. Lower scores in this dataset are because of two main reasons. First, images in this dataset are blurry and have low resolution. Second, the proposed method is sensitive to the shelf height and full visibility of the shelf. There are some images in the grocery product dataset which violate this constraint. Therefore, they led to low performance. Even with these negative effects, the proposed method was able to produce F1 scores around 0.927 with the YOLOv5x model. 

\begin{table}[htbp]
	\centering
	\caption{Object detection performance of the proposed method on the grocery product dataset.}\label{table:pla_obj_perf_test_grozi}
	\begin{tabular}{lccc}
		\toprule
        \textbf{Model}		& \textbf{Precision}	& \textbf{Recall}	& \textbf{F1 Score}	\\
		\midrule
		\textbf{Baseline}	& 0.910					& 0.935				& 0.922				\\
		\textbf{YOLOv5s}	& 0.879					& 0.895				& 0.887				\\
		\textbf{YOLOv5x}	& 0.924					& 0.929				& 0.927				\\
		\bottomrule	
	\end{tabular}
\end{table}

\subsection{Planogram Compliance Control Performance}

There is no consensus on measuring the planogram compliance control performance in literature. The closest study for this purpose is by Liu~\emph{et al.}~\citep{Liu}. Unfortunately, this study focuses on a graph theory based representation. Therefore, we provide our own measures based on precision, recall, and F1 score values. We define true positives as alignment results matching the ground truth, false positives as misalignments not in the ground truth, and false negatives as ground truth object groups missing in the alignment results.

We provide planogram compliance control performance results for the Migros dataset in Table~\ref{table:pla_compliance_perf_test_migros}. As can be seen in this table, the proposed method achieves higher performance compared to the baseline method. This is because the planogram compliance control performance is directly related to the object detection performance given in Table~\ref{table:pla_obj_perf_test_migros}. This can be expected since correctly detecting objects in the shelf image directly affects the corresponding planogram formation from the shelf.

\begin{table}[htbp]
	\centering
	\caption{Planogram compliance control performance of the proposed method on the Migros dataset.}\label{table:pla_compliance_perf_test_migros}
	\begin{tabular}{lccc}
		\toprule
        \textbf{Model}		& \textbf{Precision}	& \textbf{Recall}	& \textbf{F1 Score}		\\
		\midrule
		\textbf{Baseline}	& 0.954					& 0.980				& 0.966					\\ 
		\textbf{YOLOv5s}	& 1.000					& 1.000				& 1.000 				\\
		\textbf{YOLOv5x}	& 0.987					& 0.993				& 0.990				\\
		\bottomrule
	\end{tabular}
\end{table}

\subsection{Timing and Power Consumption Performance of the Image Acquisition and Transfer Block}

Timing performance of the image acquisition and transfer block is important since we would like to have a stand-alone system to achieve this operation. Therefore, we first analyzed timing of the methods running on the ESP-EYE module. We measured the results as 3.5 and 8.5 seconds for image acquisition and transfer operations, respectively. Based on this timing performance results and working scenario provided in Section~\ref{sec:imageaqusition}, we can report the average current drawn from the 3.3~V power supply (including wireless transfer) as 243.2~mA. Assuming that the 18650 sized LiFePO4 battery has 1500~mAh capacity, we can claim that the ESP-EYE based first section of the proposed embedded system can last approximately 24 months. This duration can be enhanced by increasing battery capacity or adding an energy harvesting module to the system. 

According to our tests, the SPV1050 solar energy harvester and $52\times27$ mm sized a-Si PV cell can extend battery life approximately 12 to 24 months depending on lighting levels 250~lux and 500~lux, respectively. Moreover, the P2110B RF energy harvester can feed 3.3~V and 0.018~mA continuously to the system under -8~dBm average input power. Hence, operation durations can be expanded further by approximately 12 more months. The harvesting options can be used as complementary to further increase the battery life. Moreover, if the lighting increase above the 1000~lux, a double sized a-Si PV cell is used or the average input RF power increases above to -3~dBm, then the system can operate infinitely without recharging or replacing the battery. Hence, the battery size became insignificant and a LiFePO4 battery with the smallest capacity can be used.

\subsection{Timing Performance of the Object Detection Block}

We analyzed timing performance of the object detection block on selected hardware platforms at maximum GPU and CPU power levels in this section. While calculating the timing performance for the object detection block, we pick the average time required to process one shelf rack image as a measure. This analysis was conducted using images from the Migros dataset. We provided the results in Table~\ref{table:timing_per}. It is important to note that if the object detection block fails to identify all objects in a shelf image, iterations in the focused and iterative search method increase, significantly impacting the timing performance. 

 \begin{table}[htbp]
 	\centering
 	\caption{Timing performance (in seconds) of the proposed method on different SBCs.}\label{table:timing_per}
 	\begin{tabular}{lrrr}
 		\toprule
         \textbf{SBCs}			& \textbf{Baseline}	& \textbf{YOLOv5s}	& \textbf{YOLOv5x}	\\
 		\midrule
 		\textbf{RPi4}			& 21.61				& 28.90				& 36.13				\\
        \textbf{Orin Nano}		& 2.70				& 5.70				& 7.43				\\
  		\textbf{AGX Orin}		& 1.86				& 3.65				& 3.76				\\
 		\textbf{PC}				& 5.96				& 6.17				& 9.65				\\
 		\bottomrule
 	\end{tabular}
 \end{table}

As can be seen in Table~\ref{table:timing_per}, YOLOv5s emerges as the fastest model across all hardware platforms despite having the lowest object detection score. Nevertheless, it cannot outperform the baseline method in terms of speed. On the other hand, YOLOv5x has the slowest timing performance, attributable to its larger model size and increased number of parameters. Therefore, the reader should take these values into account while picking the appropriate method.

NVIDIA Jetson AGX Orin stands out as the fastest SBC for all models when the hardware platforms of interest are compared. To be more specific, with GPU support, it is almost 1.7 times times faster than a PC with an Intel i7--7600U CPU while using YOLOv5s. As expected, Raspberry Pi4 is the slowest SBC due to its less powerful CPU\@. These durations imply that complete planogram compliance control in a store equipped with 100 ESP-EYE modules can take approximately 2.4, 0.5, and 0.3 hours with YOLOv5s on a single Raspberry Pi4, NVIDIA Jetson Orin Nano, and AGX, respectively. The same operations can take around 3, 0.6, and 0.3 hours with YOLOv5x on the Raspberry Pi4, NVIDIA Jetson Orin Nano, and AGX, respectively.

We should also consider the cost of mentioned SBCs in comparison. The price for the NVIDIA Jetson AGX Orin and Nano boards are USD 1999 and USD 499, respectively~\citep{orin}. The price for the Raspberry Pi4 board is USD 75~\citep{rpi}. Based on the price of the SBCs and timing values, the same tasks using YOLOv5x could be completed at cost of USD 1999, USD 998 or USD 750 using a single NVIDIA Jetson AGX Orin SBC, two NVIDIA Jetson Orin Nano SBCs or ten Raspberry Pi 4 SBCs, respectively.

\section{Final Comments}

In this study, we proposed an embedded planogram compliance control system with all its components. The methodology used in the system is the improved version of our previous work. To be more precise, the improvement has been done by adding YOLO or object detection. This way, our system precisely detects and aligns objects on shelves which is crucial for planogram compliance control accuracy. Our proposed embedded system architecture is designed for effective operation within the constraints of power consumption, cost, and physical space in retail environments. Therefore, we benefit from low-cost ESP-EYE modules for image capture and more powerful single-board computers like Raspberry Pi 4 or NVIDIA Jetson for implementation. We also applied solar and RF energy harvesting techniques to extend the battery life of our system while maintaining its compactness. This aspect is crucial for seamless integration into retail environments. Our scalability analysis, involving various hardware configurations, demonstrates the flexibility of the system in adapting to diverse performance requirements and budget constraints. Our experimental validation conducted using public datasets showcases the strengths of our approach in comparison to existing methods. The results affirm the viability of our system in smart retail applications, highlighting its scalability, efficiency, and cost-effectiveness. The next step for us is extending our method to employ machine learning and AI algorithms for better planogram alignment and matching. Also we can train lightweight models to run on low power embedded systems. Hence, it can be used in stand alone form in retail sector.

\bibliographystyle{unsrtnat}
\bibliography{planogram_control}  






\end{document}